\documentclass[10pt, a4paper]{article}

\usepackage[final]{lrec2026} % this is the new style
\usepackage{acronym}
\acrodef{eeg}[EEG]{electroencephalography}
\acrodef{erp}[ERP]{event-related potential}
\acrodef{rsvp}[RSVP]{rapid serial visual presentation}
\acrodef{spr}[SPR]{self-paced reading}
\acrodef{lm}[LM]{language model}
\acrodef{pos}[POS]{part of speech}
\acrodef{tint}[TiNT]{the Tilburg corpus of Natural Dutch Texts}

\usepackage{amsmath}
\usepackage{upgreek}

\usepackage[nohyphen]{underscore}

\usepackage{caption}
\usepackage{subcaption}
\usepackage{threeparttable}
\usepackage{tablefootnote}

\title{Modeling semantic association in self-paced reading with language model embeddings}
\name{Sara Møller Østergaard$^{\ast}$, Kenneth Enevoldsen$^{\dagger}$, Afra Alishahi$^{\ast}$,\\ {\bf \large and Bruno Nicenboim$^{\ast}$}} 

\address{
        $^{\ast}$Department of Computational Cognitive Science, Tilburg University \\
        $^{\dagger}$Center for Humanities Computing, Aarhus University \\
        s.m.ostergaard@tilburguniversity.edu\\
}

\abstract{
Semantic association between a word and its context has been identified as an important component of reading comprehension, even when word predictability is accounted for. 
Recent research has highlighted the potential of \ac{lm} embeddings to quantify semantic association. Yet, embedding-based semantic association have been operationalized in a myriad of ways.
In this study, we use embeddings from \acp{lm} to estimate semantic association on a corpus of joint \ac{eeg} and self-paced reading of natural, Dutch texts. Semantic association is calculated in ten different implementations that vary the embedding model and context lengths.
The effects of semantic association across the different implementations on the N400 and self-paced reading times are examined using Bayesian hierarchical models and Bayes factor. 
The results show that the choice of embedding model can alter the estimated effect of semantic association on both the N400 and self-paced reading times. 
Furthermore, the results demonstrate a promising potential of sentence embeddings for capturing semantic association, as only implementations relying on sentence embeddings indicate reliable results of semantic association beyond word predictability on both neural and behavioral measures.
Together, these findings highlight the importance of methodological choices in quantifying semantic association.
 \\ \newline 
}

\begin{document}

\maketitleabstract

\acresetall
\section{Introduction}

% Example from Federmeier and Kutas (1999) (Low constraint): By the end of the day, the hiker’s feet were extremely cold and wet. It was the last time he would ever buy a cheap pair of boots/sandals/jeans.
% Example from Federmeier and Kutas (1999) (high constraint): “Checkmate,” Rosaline announced with glee. She was getting to be really good at chess/ monopoly/football.

Humans process words in the context in which they are presented. How predictable a word is given its preceding context largely impacts the processing difficulty of the word \citep{kutasThirtyYearsCounting2011,ehrlichContextualEffectsWord1981, wongPredictionReadingReview2024}. 
For example, in the sentence pair, ``By the end of the day, the hiker's feet were extremely cold and wet. It was the last time he would ever buy a cheap pair of \textit{boot/jeans}.'', the final word ``boots'' is highly predictable based on the preceding context and is therefore processed more easily than the alternative ending ``jeans'', which is comparatively unpredictable in this context (example from \citealp{federmeierRoseAnyOther1999}).

Predictability of a word, or its probability given a context, has been estimated using a range of probabilistic models, including probabilistic grammars \citep{haleProbabilisticEarleyParser2001} and, more recently, next token probabilities from \acp{lm} \citep{michaelovMathematicalRelationshipContextual2024,michaelovStrongPredictionLanguage2024,frankWordEmbeddingDistance2017,xuRevisitingJokeComprehension2024, frankERPResponseAmount2015, frankEyetrackingwithEEGCoregistrationCorpus2024, pimentelEffectAnticipationReading2023}. Additionally, word predictability has been estimated using the cloze task\footnote{The cloze task is a language comprehension task in which one or more words are removed from a text and must be filled in by the participant based on contextual cues.} \citep{lukeProvoCorpusLarge2018, dambacherFrequencyPredictabilityEffects2006, bulkesSemanticConstraintReading2020}.

% maybe we could cut the surprisal theory explanation?
% An influential account of contextual processing is surprisal theory, which posits that effort required to process a word is proportional to the surprisal of the word given its context \citep{haleProbabilisticEarleyParser2001,levyExpectationbasedSyntacticComprehension2008}. Surprisal is typically formalized as an information-theoretic measure, defined as the negative log-probability of a word given its context.
% Surprisal or 
Word predictability has been able to explain important aspects of processing difficulty, however, it doesn't provide a full account. In addition to predictability, semantic association presents another factor that modulates reading comprehension \citep{kutasThirtyYearsCounting2011, brouwerGettingRealSemantic2012}. 
Semantic association refers to the degree of semantic relatedness between a target word and the context in which it is presented. While this measure is related to the predictability of the word, it has distinct properties. Using the example context from above, ``By the end of the day, the hiker's feet were extremely cold and wet. It was the last time he would ever buy a cheap pair of \textit{sandals}.'', the word ``sandals'' is unpredictable in the context, however, it is semantically associated with the context (which mentions feet). %and more so than the word ``jeans''. 
\citet{federmeierRoseAnyOther1999} show that this distinction results in different processing of these target words.

Semantic illusion has been used to study the effects of semantic association beyond word predictability. Semantic illusions refer to a phenomenon where unpredictable (or incorrect) words are temporally unnoticed because the words are semantically associated with the context. The sentence ``For breakfast the \textit{eggs} would only eat toast and jam.'', illustrates this effect, where the word ``eat'' fails to elicit the expected neural response to an unpredictable word \citep{kuperbergElectrophysiologicalDistinctionsProcessing2003}. Studies on semantic illusion report that words semantically associated with their context are processed differently (as shown with \aclu{eeg}; \acs{eeg}) compared to words that lack such associations \citep{kuperbergElectrophysiologicalDistinctionsProcessing2003, nieuwlandTestingLimitsSemantic2005, stoneRoleSyntacticSemantic2025, aurnhammerP600ContinuousIndex2023}.
Relatedly, \citet{kriegerLimitsLLMSurprisal2024} found that word predictability from \acp{lm} doesn't capture the complete role of contextual information in human sentence processing, particularly with respect to semantic association.

Processing difficulty is commonly indexed using behavioral measures such as reading times, as well as neural measures derived from \ac{eeg}, including the N400 and the P600 \ac{erp} components.
Word predictability has been shown to have robust effects on reading times and the N400 \citep{kutasThirtyYearsCounting2011, ehrlichContextualEffectsWord1981, frankERPResponseAmount2015,shainWordFrequencyPredictability2024,pimentelEffectAnticipationReading2023,frankEyetrackingwithEEGCoregistrationCorpus2024,federmeierRoseAnyOther1999}.
In contrast, semantic association between target words and their context has been investigated primarily in \ac{erp} studies, with fewer studies examining its relationship to reading times.

% Semantic association between target words and their context has mainly been studied in relation to the N400 \ac{erp} component, 
\Ac{erp} studies of semantic association have mostly focused on the N400 component, where semantic association decreases the negative amplitude of the component \citep{fischlerBrainPotentialsRelated1983, kuperbergElectrophysiologicalDistinctionsProcessing2003,federmeierRoseAnyOther1999,xuRevisitingJokeComprehension2024,broderickElectrophysiologicalCorrelatesSemantic2018,frankWordPredictabilitySemantic2017}. 
% However, evidence for semantic association effects on the N400 is mixed.
However, studies have found that the effect of semantic association on the N400 disappears when there is a delay between the semantically related context and the critical word \citep{chowWaitSecondDelayed2018, stoneRoleSyntacticSemantic2025}.
Furthermore, \citet{salicchiDifferentReadingProcessing2025} found that semantic association didn't explain variance in the N400 component when surprisal was accounted for, while it did in the P600 component, suggesting effects on later processing stages. % maybe some citations from "on the limits of llms" could also have more studies looking at the P600 component
% The effect of semantic association has also been studied in relation to reading times. 
Evidence of the effect on reading times is less explored. While some studies have found that stronger semantic association decreases reading times \citep{pynteOnlineContextualInfluences2008, mitchellSyntacticSemanticFactors2010}, other studies indicate that semantic association has no effect on reading times when excluding the variance explained by word predictability \citep{traxlerPrimingSentenceProcessing2000,frankWordEmbeddingDistance2017}.

Studies of semantic association have mostly relied on stimuli consisting of handcrafted contexts and target words, where they are either semantically similar or not \citep{federmeierRoseAnyOther1999,fischlerBrainPotentialsRelated1983,kuperbergElectrophysiologicalDistinctionsProcessing2003,stoneRoleSyntacticSemantic2025}. 
However, recent studies have attempted to estimate the semantic association using embeddings from \acp{lm} \citep{broderickElectrophysiologicalCorrelatesSemantic2018, ettingerModelingN400Amplitude2016, xuRevisitingJokeComprehension2024, michaelovStrongPredictionLanguage2024, frankWordEmbeddingDistance2017, michaelovMathematicalRelationshipContextual2024, frankWordPredictabilitySemantic2017, parvizUsingLanguageModels2011}. 
%In these approaches, embeddings of the target word and its context are extracted separately from the model, after which their similarity is computed. 
Thereby, enabling the quantification of semantic association as a continuous measure and facilitating analyses that can extend to naturalistic stimuli.

%In most studies, the similarity between the context and word embeddings is defined as the cosine similarity \citep{ ettingerModelingN400Amplitude2016, xuRevisitingJokeComprehension2024, michaelovStrongPredictionLanguage2024, frankWordEmbeddingDistance2017, michaelovMathematicalRelationshipContextual2024}. 
Embedding-based estimates of semantic association have been conceptualized in a myriad of ways. 
% mainly varying on three components:
% \begin{enumerate}
%     \item Embedding model
%     \item Calculation of context embedding
%     \item Similarity function
% \end{enumerate}
Firstly, studies deploy different embedding models for extracting the embeddings of the context and the critical word. Most studies use word embeddings, e.g., GloVe, word2vec or fastText \citep{broderickElectrophysiologicalCorrelatesSemantic2018, ettingerModelingN400Amplitude2016, xuRevisitingJokeComprehension2024, michaelovStrongPredictionLanguage2024, frankWordEmbeddingDistance2017, michaelovMathematicalRelationshipContextual2024, frankWordPredictabilitySemantic2017}, however, these models vary in model architecture, embedding size, and training data. 
Secondly, the context embedding is defined in a variety of ways. Most commonly an average of the word embeddings are used, however, which words are included in the average varies: some studies use all the words in the context \citep{michaelovMathematicalRelationshipContextual2024,michaelovStrongPredictionLanguage2024,xuRevisitingJokeComprehension2024,broderickElectrophysiologicalCorrelatesSemantic2018}, others only content words \citep{mechtenbergMeasuringBrainSensitivity2025a,frankWordPredictabilitySemantic2017} or a manually select subset of the words \citep{frankWordEmbeddingDistance2017,ettingerModelingN400Amplitude2016}. Additionally, the length of the context varies. While most studies rely on sentence-level stimuli and use all the preceding words as the context, other studies relying on stimuli consisting of longer text have defined context windows. \citet{frankWordEmbeddingDistance2017} defined the context in two separate ways: i) only the sentence preceding the critical word and ii) the four content words immediately preceding the critical word. Similarly, \citet{mechtenbergMeasuringBrainSensitivity2025a} examined local and global effects of semantic association by defining context windows of one, two, five, and ten words preceding the critical word, excluding stop words.
Finally, different functions for calculating the similarity between the embeddings of the critical word and the context have been employed: While the vast majority utilize the cosine similarity \citep{ettingerModelingN400Amplitude2016, xuRevisitingJokeComprehension2024, michaelovStrongPredictionLanguage2024, frankWordEmbeddingDistance2017, michaelovMathematicalRelationshipContextual2024}, Pearson's correlation has also been used (e.g., \citealp{broderickElectrophysiologicalCorrelatesSemantic2018}) 

% The present study investigates whether semantic association derived from \acp{lm} embeddings captures aspects of language processing not accounted for by word predictability alone. We evaluate multiple implementations of semantic association using model comparison and test their effects on self-paced reading times and the N400 \ac{erp} component. We implement semantic association utilizing embeddings from different models and define contextual windows of varying lengths.
The present study investigated whether semantic association derived from \ac{lm} embeddings captured aspects of language processing not accounted for by word predictability alone. To accommodate alternative formalizations of semantic association, we defined multiple implementations, varying the embedding model and the size of the contextual window used to compute semantic association. We evaluated these implementations using Bayesian model comparison (Bayes factor) and assess their effects on self-paced reading times and the N400 \ac{erp} component.
The results of the study showed how the choice of embedding model and the conceptualization of the context can alter the conclusions across neural and behavioral signals.

\section{Methods}

\subsection{Data}
The study used data from \aclu{tint} (\acs{tint}; \citealp{ostergaardCorpusJointEEG2025}). The corpus consists of joined recordings of \ac{eeg} and \ac{spr} from 71 participants (whereof 56 participants were included in the analysis of the current study). All participants read eight medium-length (approx. 600 words), natural, Dutch texts of different genres. Seven texts were read using a \ac{spr} paradigm, while a single text was read in a \ac{rsvp} paradigm (the exact text changing from participant to participant). In this study, we only used data recorded during \ac{spr}.  

Preprocessing of the \ac{eeg} signal and extraction of \acp{erp} were identical to that of \citet{ostergaardCorpusJointEEG2025}. Preprocessing included rereferencing of the electrodes, band-pass filtering, and artifact detection and exclusion. The N400 was defined as the mean amplitude of centroparietal electrodes in the time window 300-500ms after word onset.

\subsection{Semantic association}

Semantic association was defined as the similarity between the embedding of the context and the embedding of the critical word.
Thus, three methodological decisions were required:
(1) How to represent the embeddings of the context and the critical word, (2) what context length to use, and (3) which similarity function to apply.
% \begin{enumerate}
%     \item Embeddings of the context and the word
%     \item The context
%     \item The similarity function
% \end{enumerate}
In this paper, we defined multiple implementations of semantic association by varying the first two factors, while we used the cosine similarity as the similarity metric across all implementations. Cosine similarity was used, as it is the standard similarity measure for distributional embedding models \citep{yamada2020wikipedia2vec, reimersSentenceBERTSentenceEmbeddings2019}.

\textbf{(1) Embeddings of the context and the word:}
% embeddings
Multiple approaches exist for deriving embeddings of text using \acp{lm}. Embeddings can be uncontextualized, such as, GloVe, word2vec, or fastText \citep{penningtonGloVeGlobalVectors2014, mikolovEfficientEstimationWord2013, bojanowskiEnrichingWordVectors2017}. Such models produce a single embedding for each word in isolation. Alternatively, embeddings can be contextualized. Contextualized embeddings can be derived from transformer models, including both encoders such as BERT \cite{devlinBERTPretrainingDeep2019} and generative models such as GPT and LLama \cite{gpt, llama} by retrieving embeddings from the last hidden state. However, embeddings derived directly from pre-trained models typically perform poorly, and thus it has become the norm to adapt contextualized transformer models for embedding tasks, such as semantic text similarity \citep{reimersSentenceBERTSentenceEmbeddings2019, gaoSimCSESimpleContrastive2021, liTransformingGenericCoder2025}. 
% In this study, we used one contextualized and one uncontextualized embedding model to extract semantic association. 

An initial exploration of implementations of semantic association using different embedding models was conducted with simple  sentences where the differences in semantic association were handcrafted. 
The results of the exploration indicated that both contextualized and uncontextualized embedding models were able to differentiate words semantically associated with the context from unrelated words. The results from the models were similar within embedding type (i.e., contextualized or uncontextualized).\footnote{Results of initial exploration can be found in appendices.}
% Noteably, sentence embeddings provided a clearer delineation, highlighting the potential of this novel approach for computing semantic association.
As such, we selected two candidate models: an uncontextualized word embedding model and a contextualized sentence embedding model.\footnote{For historical reasons, we call these sentence embeddings, as they initially were trained to embed sentences. However, they have since been expanded to embed entire documents.}

For the uncontextualized model, we used the word2vec model \texttt{wikipedia2vec\_nlwiki\_20180420\_300d}\footnote{Model revisions can be found in appendices.} \citep{yamada2020wikipedia2vec}. This model was chosen as the training procedure matched previous literature \citep{broderickElectrophysiologicalCorrelatesSemantic2018, ettingerModelingN400Amplitude2016, frankWordEmbeddingDistance2017, frankWordPredictabilitySemantic2017} and its training data overlaps with the \acs{tint} corpus \citep{ostergaardCorpusJointEEG2025}.
As the model only produces one embedding for each word, we used two methods for obtaining the embedding of the context: i) the average of the embeddings of all the words in the context (denoted \textit{WE}), and ii) the average of all the content words in the context (denoted \textit{CWE}). We used the \textit{nl\_core\_news\_sm} model from \texttt{spaCy} to extract the \ac{pos} tags. Content words were identified as words with the \ac{pos} tags noun, verb, adjective, or adverb.

For the sentence embedding we used \texttt{e5-large-trm-nl} \citep{banar2025mtebnle5nlembeddingbenchmark} as it performed well on the Dutch embedding benchmark (MTEB(nld, v1); \citealp{banar2025mtebnle5nlembeddingbenchmark, enevoldsen2025mmtebmassivemultilingualtext}). 
Sentence embeddings are trained to produce an aggregated embedding over multiple words, thus, it didn't require post-hoc averaging to obtain the embedding of the context. Implementations with sentence embeddings are denoted \textit{SE}.

\textbf{(2) Context length:}
% Context lengths
Contexts of varying length were defined to examine local and global effects of semantic association. Four distinct context lengths were used. 
First, a naive context consisting of all words preceding the critical word was used (\textit{All}). Second, we defined a context consisting of all words in the preceding sentence, as well as in the sentence to which the critical word belonged (\textit{Sentence(N=1)}). Finally, we defined a windowed context, where the context consisted of a fixed number of content words before the target. Here, we used windows of one and two (\textit{Windowed(N=1)} and \textit{Windowed(N=2)}). The windowed implementation was only defined with (content) word embeddings. 

In addition to the contexts of different lengths, we defined a weighted average of the word embeddings. The weights followed an exponential forgetting curve, thus, assuming words appearing closer to the critical word were more important. This was implemented as in Equation \ref{eq:we weighted}:

\begin{equation}
    \text{WE, Weighted} = \sum_{i = 1}^N  2^{\frac{-i}{4}} \cdot \text{similarity}(w_c, w_i)
    \label{eq:we weighted}
\end{equation}
Here, $w_c$ is the word embedding of the critical word and $w_i$ the word embedding of the word $i$ words away from the critical word. The equation sums over all words preceding the critical word. The denominator (4) determines the half-life of the decay and was chosen such that words at a distance of ten or more from the critical word receive minimal weights. As for the other implementations, the similarity was calculated with the cosine similarity. The weighted average was implemented with word embeddings of all words and word embeddings of content words only.
All the implementations of semantic association are summarized in Table \ref{tab:implementations}. \\

\begin{table*}[htb]
    \centering
    \begin{tabular}{llp{0.3\textwidth}l}
    \hline
    \textbf{Name} & \textbf{Embedding model} & \textbf{Context} & \textbf{Words} \\ 
    \hline
    SE, All & Sentence Embeddings & All preceding words & All \\
    SE, Sentence(N=1) & Sentence Embeddings & One sentence before the target sentence & All \\
    WE, All & Word Embeddings & All preceding words & All \\
    WE, Sentence(N=1) & Word Embeddings & One sentence before the target sentence & All \\
    WE, Weighted & Word Embeddings & All preceding words (weighted) & All \\
    CWE, All & Word Embeddings & All preceding words & Content words \\
    CWE, Sentence(N=1) & Word Embeddings & One sentence before the target sentence & Content words \\
    CWE, Weighted & Word Embeddings & All preceding words (weighted) & Content words \\
    CWE, Windowed(N=1) & Word Embeddings & One content word preceding the target  & Content words \\
    CWE, Windowed(N=2) & Word Embeddings & Two content word preceding the target  & Content words \\
    \hline
    \end{tabular}
    \caption{All implementations of semantic association used in the current study. The implementations will be referred to by the name in the name column.}
    \label{tab:implementations}
\end{table*}

\begin{figure}[htb]
    \centering
    \includegraphics[width=.5\textwidth]{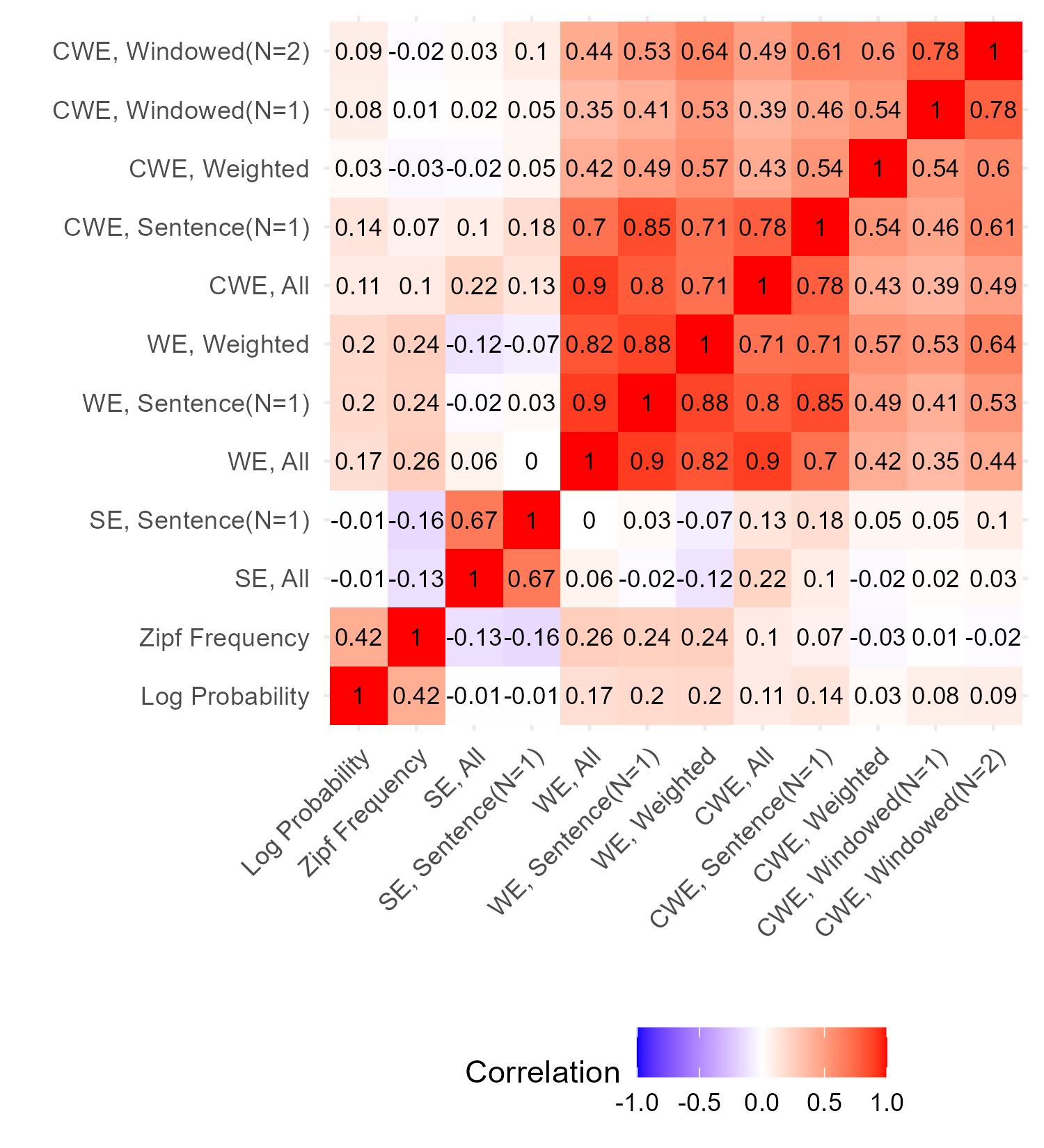}
    \caption{Pearson's correlation coefficients between implementations of semantic association, log-probability of words, and Zipf word frequency for all content words in the corpus.}
    \label{fig:correlations}
\end{figure}

Correlations between semantic association for content words in the corpus extracted using the different implementations are shown in Figure \ref{fig:correlations}. Implementations based on the same type of embedding (i.e., word or sentence embeddings) are strongly correlated, suggesting that they index similar sources of variance. In contrast, correlations across embedding types are substantially weaker.

\subsection{Regression models and model comparison}

Bayesian hierarchical models were fitted to examine the effect of the different implementations of semantic association on the two dependent variables: self-paced reading times and the N400. The models were fitted in Stan (version 2.32.2; \citealp{stan}) using the \texttt{brms} package (version 2.22.0; \citealp{brms}) in R \citep{rcoreteam}. All predictors were z-score standardized. Words with reading times lower than 100 ms or greater than 3000 ms were excluded from analysis. Only content words (i.e., nouns, verbs, adjectives, and adverbs) were included in the analysis. As embeddings extracted from the word2vec model only exist for a finite number of words, the data loss slightly differed across different implementations of semantic association.\footnote{Data loss across all implementations is reported in appendices.} The models were fitted on complete cases across all implementations.

The models were run with two predictors, log-probability $lp$, estimated by the average word probability from four GPT models (see \citealp{ostergaardCorpusJointEEG2025}), and semantic association $sem$. One regression model for each of the two dependent variables and each of the 10 implementations of semantic association was fitted, resulting in 20 separate models. Uncorrelated group-level intercepts and slopes for both $lp$ and $sem$ were estimated for each participant, document, and word.
For the reading time (RT) model, a log-normal likelihood was used, while for the N400 model, a Gaussian likelihood was used (See Equations \ref{eq:RT_model}, \ref{eq:ERP_model}, and \ref{eq:semantic_association}).

\begin{align}
    \text{RT} &\sim LogNormal(\mu, \sigma) \label{eq:RT_model} \\
    \text{N400} &\sim Normal(\mu, \sigma) \label{eq:ERP_model}
\end{align}
\begin{align}
    \mu = &\alpha + u_{participant,0} + u_{document,0} + \nonumber \\
    &u_{word,0} + (\beta_1 + u_{participant,1} +\label{eq:semantic_association} \\
    &u_{document,1} + u_{word,1}) \cdot lp + (\beta_2 +  \nonumber \\
    &u_{participant,2} + u_{document,2} + u_{word,2}) \cdot sem \nonumber
\end{align}

Different priors were used for the reading times and the N400 model, as the scales of the dependent variables were different, i.e., reading times in ms and \ac{erp} components in $\upmu$V. For all models, regularizing priors were used to ensure stable and plausible estimates \citep{nicenboimComputationalBayesianData2025}. The priors for the reading times model were as follows:
\begin{align*}
    \alpha &\sim Normal(5.5,1) \\
    \beta &\sim Normal(0, .1) \\
    u &\sim Normal(0, sd) \\
    sd &\sim Normal_+(0, .5) \\
    \sigma &\sim Normal_+(0, .5)
\end{align*}
The priors for the models of the \ac{erp} components were:
\begin{align*}
    \alpha &\sim Normal(0,20) \\
    \beta &\sim Normal(0, 10) \\
    u &\sim Normal(0, sd) \\
    sd &\sim Normal_+(0, 10) \\
    \sigma &\sim Normal_+(0, 10)
\end{align*}

To assess the influence of the various implementations of semantic association on reading comprehension (i.e., reading times and the N400), we used Bayes factors. Bayes factor provides a framework for Bayesian hypothesis testing by quantifying evidence in favor of a model ($M_0$) given another ($M_1$). This is calculated as the ratio between the marginal likelihoods of the two models, which in turn responds to two hypotheses.  (see Equation \ref{eq:bf}).
\begin{equation}
    BF_{01} = \frac{p(y | M_0)}{p(y | M_1)}
    \label{eq:bf}
\end{equation}

As such, a Bayes factor of one indicates no evidence for either model, a Bayes factor of 10 would be strong evidence for $M_0$, and a Bayes factor of $1/10$ indicates strong evidence for $M_1$.
We used the Savage-Dickey density ratio method to calculate Bayes factor, as it provides a convenient method for computing Bayes factor for nested models with a point null hypothesis \citep{dickeyWeightedLikelihoodRatio1970}. We specifically tested the null hypothesis that there's no effect of semantic association in the models when log-probability is included. The Savage-Dickey ratio was calculated separately for each model as in Equation \ref{eq:savage-dickey ratio}.

\begin{equation}
    BF_{01} = \frac{p(\beta_2 = 0| y)}{p(\beta_2 = 0)}
    \label{eq:savage-dickey ratio}
\end{equation}

Here, $y$ denotes the observed data and $\beta_2$ is the coefficient for semantic association. 

As Bayes factor is sensitive to the choice of prior, we conducted a sensitivity analysis by varying the width of the prior for $\beta_2$ while keeping the priors for all other parameters fixed. For the reading times models, we used additional priors of $\beta_2 \sim Normal(0, .05)$ and $\beta_2 \sim Normal(0, .5)$, and for the N400 models, $\beta_2 \sim Normal(0, 1)$ and $\beta_2 \sim Normal(0, 2)$. For a more elaborate explanation of Bayes factor and the Savage-Dickey ratio, see \citet{nicenboimStatisticalMethodsLinguistic2016}.

Most models were fitted using four chains with 2,000 iterations, where half the iterations were warm-up samples. However, six models required 3,000 iterations to ensure stable posterior sampling.
The models reported in this paper had no divergent transitions, $\hat{R}s \leq 1.03$, and the number of bulk and tail effective samples was at least 119 and an average of 1,477.

\section{Results}

\begin{figure}[htb]
    \centering
    \includegraphics[width=.5\textwidth]{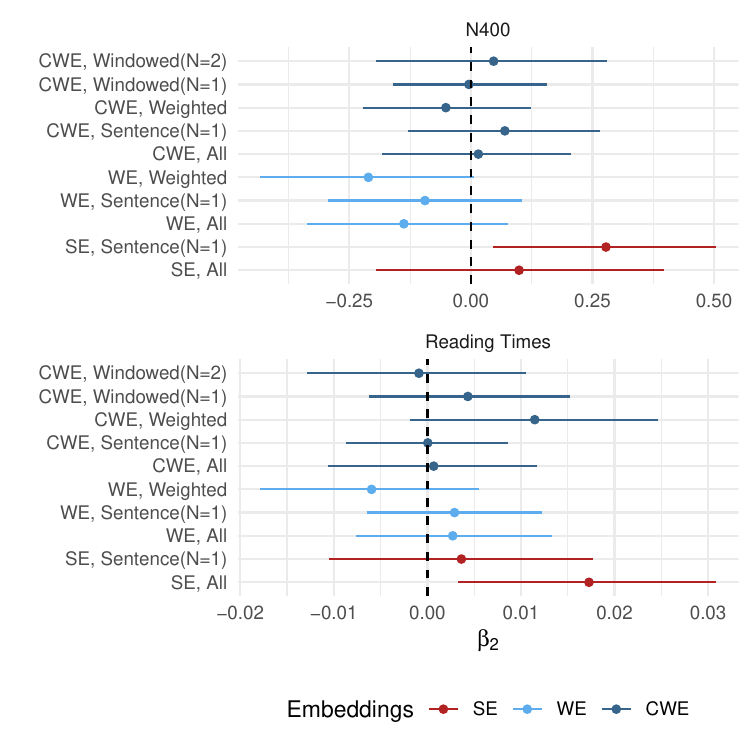}
    \caption{Regression coefficients and 95\% credible intervals for semantic association $\beta_2$ as estimated by the different implementations. Every point represents $\beta_2$ from a separate regression model. The models with N400 as the dependent variable are measured in $\upmu$V, while the reading times models are measured in ms (thus, the scales of the x-axis differ across the two).}
    \label{fig:regression coefficients}
\end{figure}

\begin{figure}[htb]
    \centering
    \includegraphics[width=.5\textwidth]{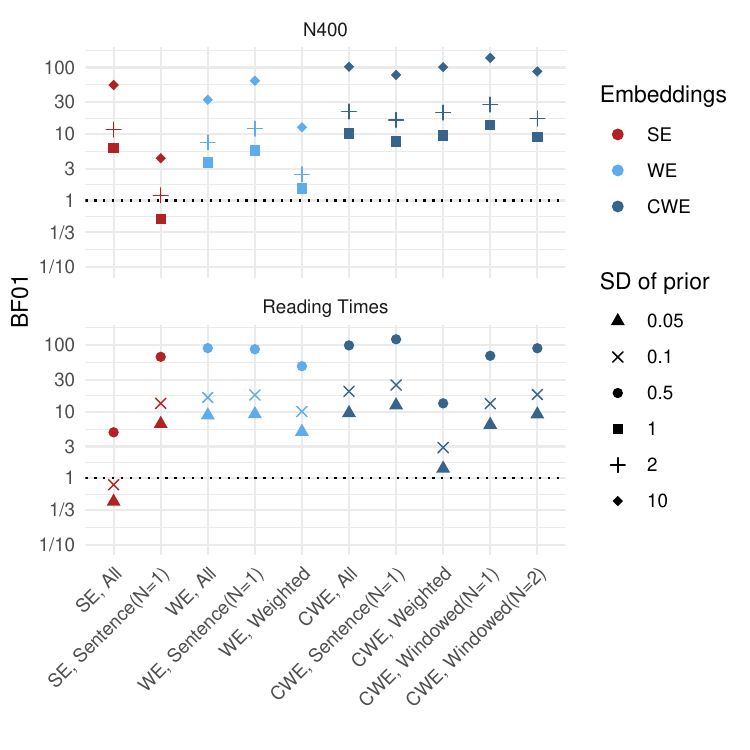}
    \caption{Bayes factor. $BF01 > 1$ indicates more evidence for the null hypothesis (i.e., no effect of semantic association) and $BF01 < 1$ indicates more evidence for the alternative hypothesis (i.e., an effect of semantic association). Each Bayes factor was calculated for separately fitted models with different standard deviations (SD) for the prior of $\beta_2$.}
    \label{fig:bf}
\end{figure}

Figure \ref{fig:regression coefficients} displays the coefficients for semantic association ($\beta_2$ in Equation \ref{eq:semantic_association}) estimated by the 20 regression models using the values from the different implementations of semantic association to predict the N400 and self-paced reading times. Bayes factor for $\beta_2$ across the regression models reported in Figure \ref{fig:bf}.
The results from the Bayes factor showed anecdotal evidence ($BF01 \in \{1, 1/3\}$) for an effect of semantic association in only two models (\textit{SE, Sentence(N=1)} for the N400 and \textit{SE, All} for reading times). Across the rest of the models for both dependent variables, there was the most evidence for the null hypothesis, i.e., no effect of semantic association.

\textbf{Embedding models:} The results of the regression models indicate that the choice of embedding model when calculating semantic association impacts the estimated effects on neural and behavioral measures. This pattern was particularly pronounced for the models of the N400. The estimated effect of semantic association on the N400 when using sentence embeddings (\textit{SE}) was positive, meaning that less semantically associated words elicited a more negative N400 amplitude, consistent with previous literature \citep{fischlerBrainPotentialsRelated1983, kuperbergElectrophysiologicalDistinctionsProcessing2003,federmeierRoseAnyOther1999,xuRevisitingJokeComprehension2024,broderickElectrophysiologicalCorrelatesSemantic2018,frankWordPredictabilitySemantic2017}. 
% This direction would be expected following previous literature on the N400 and semantic association \citep{}.
In contrast, when semantic association was calculated using word embeddings (\textit{WE}), the direction of the effect reversed, i.e., a negative estimate.
% For the models using word embeddings to calculate semantic association (\textit{WE}), the direction of the effect was in the opposite direction. 
When semantic association was calculated using the same word embeddings but only embeddings of the content words in the context (\textit{CWE}), the estimated effect of semantic association was close to zero.
For reading times, only the model of semantic association from the \textit{SE, All} implementation indicated an effect. This model estimated a positive effect of semantic association on reading times; thus, reading times increased when words were more semantically associated to the context.
The estimated coefficients for semantic association for the rest of the models were smaller and generally close to zero.

\textbf{Context length:} The results show that the length of the context matters only for the semantic association defined with sentence embeddings. The implementations of semantic association using word embeddings (both \textit{WE} and \textit{CWE}) showed similar effects on both the N400 and reading times across all contexts (\textit{All}, \textit{Sentence}, \textit{Weighted}, and \textit{Windowed}). For the implementations relying on sentence embeddings, the effect of context appeared to play a more substantial role. On the N400, the effect of semantic association was largest for the regression model with \textit{SE, Sentence(N=1)}, while the largest effect of semantic association on reading times was estimated by the model with \textit{SE, All}.
% Note that these were also the models showing weak evidence for effects in the Bayes factor analysis.

\section{Discussion} 

% \begin{itemize}
%     \item Our results say: There's a difference in effects when using different embedding types
%     \item Other studies have used word embeddings - however, our results indicate promising results of using underexplored sentence embeddings
%     \item Why would they be different? Could it be the averaging of multiple word embeddings (which have not been trained for this)? Maybe longer contexts play a role - that's why we don't replicate results from xyz.
%     \item Global/local context: does it matter or not? For sentence embeddings we have a shift in which on predicts better + see example sentence. For word embeddings [...]. It is hard to say anything conclusive from these results about this.
% \end{itemize}

\begin{figure*}[htb]
    \centering
    \includegraphics[width=0.99\textwidth]{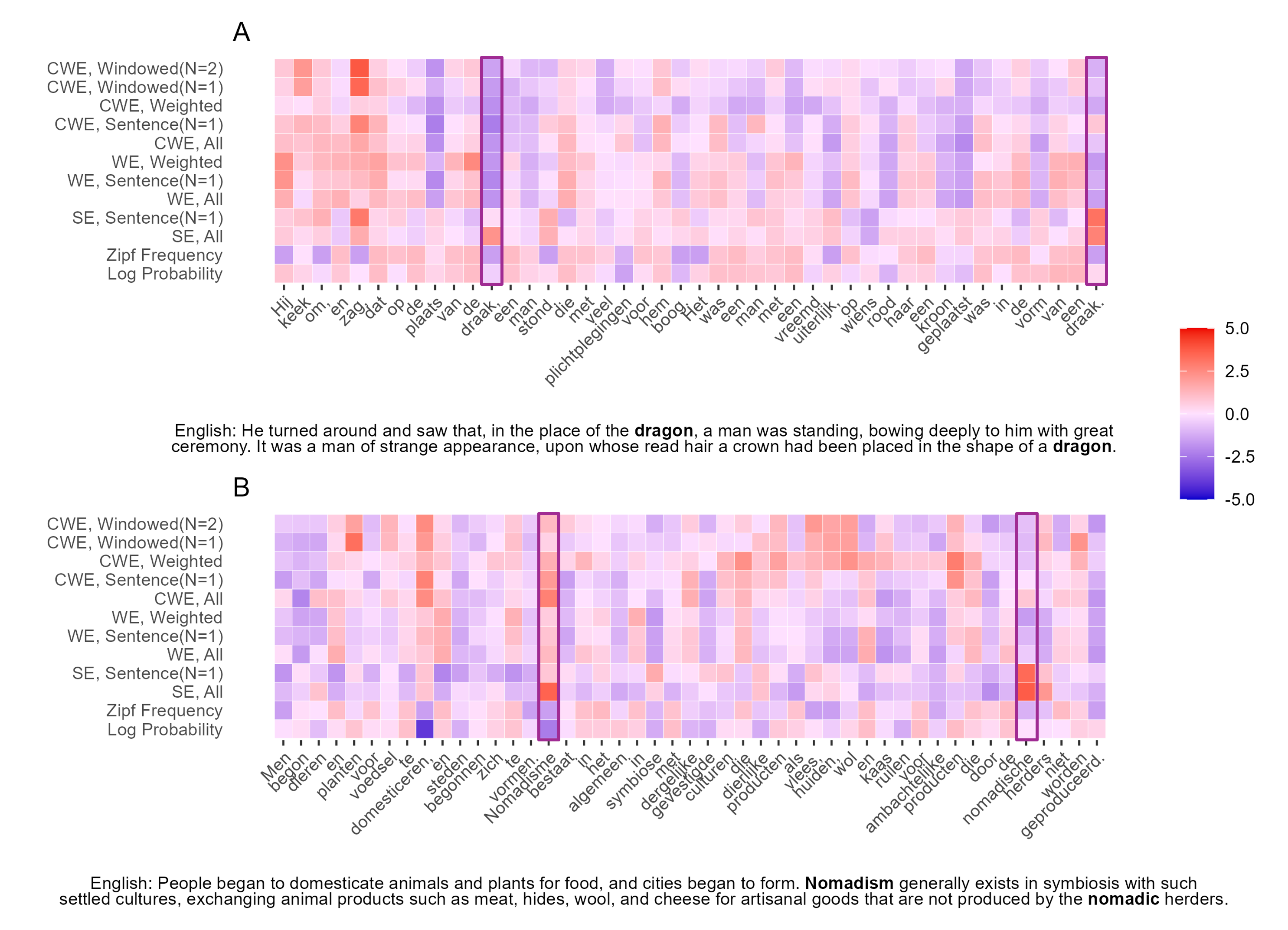}
    \caption{Semantic association (as estimated by different implementations), log-probability, and Zipf frequency of words in two sentence pairs from two different documents in the corpus. Highlighted are the words ``draak'' (English: ``dragon'') in sentences A and ``Nomadisme'' (English: ``Nomadism'') and ``nomadische'' (English: ``nomadic'') in sentences B. All variables (i.e., log probability, Zipf frequency, and semantic associations) are z-score standardized.}
    \label{fig:example sentences}
\end{figure*}

In this study, we employed both uncontextualized word embeddings and contextualized sentence embeddings to estimate semantic association. These two embedding types appear to capture distinct patterns in the text. This is both apparent from the correlations between the different implementations of semantic association (Figure \ref{fig:correlations}) but, more importantly, for the estimated effects of semantic association on self-paced reading times and the N400 too (Figure \ref{fig:regression coefficients} and Figure \ref{fig:bf}).
The results of the regression models show that the type of embeddings used influences the estimated effect of semantic association. This finding is most pronounced for the N400, where the estimated effect reverses direction depending on whether semantic association is computed using sentence or word embeddings. A positive effect of semantic association on the N400 is found when using sentence embeddings, while the opposite (i.e., a negative effect) is estimated with word embedding-based semantic association. 
In contrast, previous literature using similar uncontextualized word embeddings to calculate semantic association reports a positive effect of semantic association on the N400 \citep{broderickElectrophysiologicalCorrelatesSemantic2018,frankWordPredictabilitySemantic2017,xuRevisitingJokeComprehension2024}.
It is important to note that the Bayes factor indicated no evidence for the negative effects estimated for word embedding-based semantic association, however, anecdotal evidence for one of the models with a positive effect for semantic association from sentence embeddings.

What could be possible explanations for the observed difference in semantic association when computed with different types of embedding models? One important distinction between sentence embeddings and word embeddings when using them to create context representations lies in how information was aggregated.
Although sentence embeddings output an aggregation of embeddings too, the model has been trained to produce a semantically coherent representation in which more informative words receive greater weight. In this study, the implementations of semantic association using word embeddings from word2vec relied on a naive approach, where an unweighted average was utilized, inspired by previous approaches \citep{michaelovMathematicalRelationshipContextual2024,michaelovStrongPredictionLanguage2024,xuRevisitingJokeComprehension2024,broderickElectrophysiologicalCorrelatesSemantic2018, mechtenbergMeasuringBrainSensitivity2025a,frankWordPredictabilitySemantic2017,frankWordEmbeddingDistance2017,ettingerModelingN400Amplitude2016}. As such, important information could be lost in the context representations derived from the word embedding implementations — especially for the implementations of longer contexts (i.e., \textit{WE, All} and \textit{WE, Sentence(N=1)}). Studies finding positive effects of word embedding-based semantic association on the N400 have generally relied on shorter contexts (either because sentence-level stimuli were used or because they defined short context lengths), thus minimizing the information loss when averaging over embeddings. This speculation is supported
% by the absence of the negative effect in the windowed models (\textit{CWE, Windowed(N=1)} and \textit{CWE, Windowed(N=2)}), as well as 
by our initial exploration, where both sentence and word embeddings produced effects in the same direction using sentence pairs from \citet{federmeierRoseAnyOther1999}.\footnote{Results of initial exploration can be found in appendices} 
While the weighted implementations of semantic association (\textit{WE, Weighted} and \textit{CWE, Weighted}) were cognitively motivated implementations, discounting the influence of word embeddings on the overall average based an exponential forgetting curve, this approach didn't seem promising given the results of the current study. The weights were solely based on distances to the critical word, thus words were not weighted based on their semantic relevance.
% The weighted implementations of semantic association (\textit{WE, Weighted} and \textit{CWE, Weighted}) served as a cognitively motivated average, discounting the influence of word embeddings on the overall average based an exponential forgetting curve. The weights were solely based on distances to the critical word, thus words were not weighted based on their semantic relevance. However, this approach didn't seem promising given the results of the current study.
% While the weighted implementations of semantic association (\textit{WE, Weighted} and \textit{CWE, Weighted}) were cognitively motivated implementations, discounting the influence of word embeddings on the overall average based an exponential forgetting curve, this approach didn't seem promising given the results of the current study. 
% The weighted model only took the distance to the critical word into account, thus, words were not weighted based on how informative they were.

To our knowledge, no previous works have used sentence embeddings for studying semantic association in sentence processing. 
The present findings suggest that this approach provides a promising method for estimating semantic relations between contexts and target words. Implementations based on sentence embeddings showed the most reliable effects on both the N400 and self-paced reading times, as reflected by the size of the regression coefficients and Bayes factors.
Post-hoc qualitative analyses further indicated that sentence embeddings are more sensitive to the general themes of the texts compared to averaged word embeddings.
Figure \ref{fig:example sentences} illustrates a difference between sentence embeddings and word embeddings for calculating semantic association in two examples of sentence pairs from the corpus.
In the first pair, the word ``dragon'' appears in both sentences. Only the \textit{SE, All} implementation captures the association between ``draak'' (English: ``dragon'') and the story ``Mijn Heer Zak met Rijst''\footnote{A fairy tale about a Dragon King.} in the first sentence, while \textit{SE, Sentence (N=1)} detects the association when the word reappears in the second sentence. In contrast, none of the word embedding implementations indicate a strong association for ``dragon'' in either instance. In sentence pair B from the text ``Nomadisch pastoralisme'', a similar pattern is observed with the two words ``Nomadisme'' (English: ``Nomadism'') and ``nomadische'' (English: ``nomadic''). While these examples were selected for illustrative purposes, they suggest that sentence embeddings capture a thematically coherent representation of semantic association in natural texts. 

% what about context?
% what about the direction of the effect on reading times?

Naturally, this interpretation depends on the operationalization of semantic association. In the present study, semantic association was defined as the similarity between embeddings of the context and the critical word, following prior work \citep{broderickElectrophysiologicalCorrelatesSemantic2018, ettingerModelingN400Amplitude2016, xuRevisitingJokeComprehension2024, michaelovStrongPredictionLanguage2024, frankWordEmbeddingDistance2017, michaelovMathematicalRelationshipContextual2024, frankWordPredictabilitySemantic2017}. This operationalization assumes that the embeddings encode multiple aspects of meaning, including both shared features (e.g., \textit{nurse} and \textit{mechanic} as occupations) and thematic relations (e.g., \textit{nurse} and \textit{hospital} as related). Consequently, embedding similarity captures similarities in the features of the word as well as their relatedness. Following this definition, repeated words will inflate semantic association, as the similarity between two identical embeddings is one (i.e., maximum semantic association). However, this property applies for all the implementations of semantic association considered in the current work, thus, the effect of repetition can't account for the differences observed in example A in Figure \ref{fig:example sentences}.
\\

% Limitations/future studies:
% \begin{itemize}
%     \item We only use one study (with longer texts in Dutch) - these methods should be (more thoroughly) validated/estimated on other corpora, e.g., to see if it is possible to replicate results from previous studies with both embedding models.
%     \item Choice of embeddings could play a role - and was not thoroughly studied here. Dimension of embeddings (especially for the word embeddings - where we naively average). Also, it could make sense to use embeddings of same size and same training accross the two types of implementations (i.e., SE and WE). E.g., by extracting the uncontextualized embedding layer of the sentence embedding models.
%     \item Maybe the embedding models could perform worse on Dutch as compared to English.
%     \item Maybe the effect of semantic association in dependent on predictability rather than additive. I.e., maybe people only pay attention to predictability when there's a low semantic association (e.g., ``A plane crashed between the border of Canada and America. Where do you bury the survivors?'')
% \end{itemize}

The results of the current study are exploratory, and further work is required to identify under which conditions specific implementations of \ac{lm} embedding-based estimates of semantic association differ. The analysis was based exclusively on texts from a single corpus (\ac{tint}; \citealp{ostergaardCorpusJointEEG2025}), which consists of medium-length, Dutch texts. Not only does this corpus stand in contrast to previously used stimuli by the length of the texts (as touched upon above), but also in the language. As Dutch has been less extensively studied than English, the quality of the embedding models may differ, potentially affecting their performance. As such, semantic association as estimated by the different implementations in this study should be validated on other corpora to determine whether it is possible to replicate previously reported effects of semantic association. 

The most prominent finding of this study lies in the importance of the embedding model for estimating semantic association. Only one word embedding model and one sentence embedding model were included in the analysis, as initial explorations indicated minimal differences between models within each embedding type.
% During initial exploration, other embedding models were considered, and here we found no immediate difference for different word embedding or sentence embedding models. For this reason, we continued the analysis with just one of each. 
However, in light of the results of the current study, further exploration of different embedding models would be interesting. 
% Dimension of embeddings (especially for the word embeddings - where we naively average). Also, it could make sense to use embeddings of same size and same training accross the two types of implementations (i.e., SE and WE). E.g., by extracting the uncontextualized embedding layer of the sentence embedding models.

% Finally, in this study, the effect of semantic association on reading comprehension is assumed to be additive to word predictability. However, predictability and semantic association might be interacting, i.e., people might only pay attention to predictability for words with low semantic association. This could explain why people don't notice the unpredictability of the word ``survivors'' in the sentence ``When an aircraft crashes, where should the survivors be buried?'' for \citet{bartonCaseStudyAnomaly1993}.

%%% Add to discussion after reviews %%%
% Relatedness vs similarity: The embedding vectors are assumed to contains multiple sources of information - here among information indicating features of the entity (e.g., nurse and mechanic both refers to a job) and thematic properties (e.g., egg and breakfast). 
% Repetition: In the example of the dragon one could argue that the effect is simply one of repetition. However, we would expect this method (using embeddings) would generate high semantic association for repetition has embedding(word) = embedding(w), thus, semantic association(word, word) = 1 

\section{Conclusion}
This study examined the effects of \ac{lm} embedding-based semantic association on the self-paced reading of medium-length, Dutch texts. The findings demonstrate that the conclusions critically depend on how semantic association is implemented, particularly with respect to the embedding model.
While uncontextualized word embeddings (e.g., word2vec) have previously been used to examine semantic association in sentence processing and showed effects on the N400 \citep{xuRevisitingJokeComprehension2024,broderickElectrophysiologicalCorrelatesSemantic2018,frankWordPredictabilitySemantic2017}, we observed no effects on either the N400 or self-paced reading times. In contrast, semantic association estimated with sentence embeddings was found to be predictive of processing difficulty. These results suggest sentence embeddings to be a promising approach for examining the effects of semantic association in natural reading.

% However, in the \ac{tint} data, sentence embeddings perform noticeably better than word embeddings - to the extent where word embeddings don't even seem to capture the overarching theme of the text (example with dragon or nomads - see supervision slides). Could it be that word embeddings work well in short and manually curated contexts (where the semantic association between words is very clear), but less so in natural texts? It might be that sentence embeddings perform better, as they weight the importance of the words in the context more reliably in accordance with the overarching theme of the text.

\section{Code availability}
Code for reproducing the analysis is publicly available from GitHub at \url{https://github.com/saraoe/semantic_association}.

\section{Ethics Statement}
The study utilized human data from \acf{tint} collected by \citet{ostergaardCorpusJointEEG2025}. The data has received an ethics approval and is licensed under a CC-BY-NC-SA license. 

% \section{Acknowledgements}

\section{Bibliographical References}

\bibliographystyle{lrec2026-natbib}
\bibliography{refs}

@article{federmeierRoseAnyOther1999,
    title      = {A {{Rose}} by {{Any Other Name}}: {{Long-Term Memory Structure}} and {{Sentence Processing}}},
    shorttitle = {A {{Rose}} by {{Any Other Name}}},
    author     = {Federmeier, Kara D. and Kutas, Marta},
    year       = 1999,
    month      = nov,
    journal    = {Journal of Memory and Language},
    volume     = {41},
    number     = {4},
    pages      = {469--495},
    issn       = {0749596X},
    doi        = {10.1006/jmla.1999.2660},
    urldate    = {2025-12-06},
    copyright  = {https://www.elsevier.com/tdm/userlicense/1.0/},
    langid     = {english}
}

@misc{mechtenbergMeasuringBrainSensitivity2025a,
    title     = {Measuring Brain Sensitivity to Semantic Distance in Spoken Narrative Comprehension},
    author    = {Mechtenberg, Hannah and Reilly, James and Myers, Emily B and Peelle, Jonathan E},
    year      = 2025,
    month     = jun,
    number    = {dtnjm\_v1},
    publisher = {PsyArXiv},
    doi       = {10.31234/osf.io/dtnjm_v1},
    urldate   = {2025-11-28}
}

@article{michaelovMathematicalRelationshipContextual2024,
    title    = {On the {{Mathematical Relationship Between Contextual Probability}} and {{N400 Amplitude}}},
    author   = {Michaelov, James A. and Bergen, Benjamin K.},
    year     = 2024,
    month    = jun,
    journal  = {Open Mind},
    volume   = {8},
    pages    = {859--897},
    issn     = {2470-2986},
    doi      = {10.1162/opmi_a_00150},
    urldate  = {2025-12-15}
}

@article{michaelovStrongPredictionLanguage2024,
    title      = {Strong {{Prediction}}: {{Language Model Surprisal Explains Multiple N400 Effects}}},
    shorttitle = {Strong {{Prediction}}},
    author     = {Michaelov, James A. and Bardolph, Megan D. and Van Petten, Cyma K. and Bergen, Benjamin K. and Coulson, Seana},
    year       = 2024,
    month      = apr,
    journal    = {Neurobiology of Language},
    volume     = {5},
    number     = {1},
    pages      = {107--135},
    issn       = {2641-4368},
    doi        = {10.1162/nol_a_00105},
    urldate    = {2025-11-28}
}

@article{xuRevisitingJokeComprehension2024,
    title      = {Revisiting {{Joke Comprehension}} with {{Surprisal}} and {{Contextual Similarity}}: {{Implication}} from {{N400}} and {{P600 Components}}},
    shorttitle = {Revisiting {{Joke Comprehension}} with {{Surprisal}} and {{Contextual Similarity}}},
    author     = {Xu, Haoyin and Nakanishi, Masaki and Coulson, Seana},
    year       = 2024,
    journal    = {Proceedings of the Annual Meeting of the Cognitive Science Society},
    volume     = {46},
    number     = {0},
    urldate    = {2025-11-28},
    langid     = {english}
}

@misc{reimersSentenceBERTSentenceEmbeddings2019,
    title         = {Sentence-{{BERT}}: {{Sentence Embeddings}} Using {{Siamese BERT-Networks}}},
    shorttitle    = {Sentence-{{BERT}}},
    author        = {Reimers, Nils and Gurevych, Iryna},
    year          = 2019,
    month         = aug,
    number        = {arXiv:1908.10084},
    eprint        = {1908.10084},
    primaryclass  = {cs},
    publisher     = {arXiv},
    doi           = {10.48550/arXiv.1908.10084},
    urldate       = {2025-12-15},
    archiveprefix = {arXiv}
}

@article{broderickElectrophysiologicalCorrelatesSemantic2018,
    title     = {Electrophysiological {{Correlates}} of {{Semantic Dissimilarity Reflect}} the {{Comprehension}} of {{Natural}}, {{Narrative Speech}}},
    author    = {Broderick, Michael P. and Anderson, Andrew J. and Liberto, Giovanni M. Di and Crosse, Michael J. and Lalor, Edmund C.},
    year      = 2018,
    month     = mar,
    journal   = {Current Biology},
    volume    = {28},
    number    = {5},
    pages     = {803-809.e3},
    publisher = {Elsevier},
    issn      = {0960-9822},
    doi       = {10.1016/j.cub.2018.01.080},
    urldate   = {2025-10-16},
    langid    = {english},
    pmid      = {29478856}
}

@article{frankWordEmbeddingDistance2017,
    title    = {Word {{Embedding Distance Does}} Not {{Predict Word Reading Time}}},
    author   = {Frank, Stefan L.},
    year     = 2017,
    journal  = {Proceedings of the Annual Meeting of the Cognitive Science Society},
    volume   = {39},
    number   = {0},
    urldate  = {2025-12-03},
    langid   = {english}
}

@article{ettingerModelingN400Amplitude2016,
    title    = {Modeling {{N400}} Amplitude Using Vector Space Models of Word Representation},
    author   = {Ettinger, Allyson and Feldman, Naomi and Resnik, Philip and Phillips, Colin},
    year     = 2016,
    journal  = {Proceedings of the Annual Meeting of the Cognitive Science Society},
    volume   = {38},
    number   = {0},
    urldate  = {2025-11-27},
    langid   = {english}
}

@article{kutasThirtyYearsCounting2011,
    title      = {Thirty {{Years}} and {{Counting}}: {{Finding Meaning}} in the {{N400 Component}} of the {{Event-Related Brain Potential}} ({{ERP}})},
    shorttitle = {Thirty {{Years}} and {{Counting}}},
    author     = {Kutas, Marta and Federmeier, Kara D.},
    year       = 2011,
    month      = jan,
    journal    = {Annual Review of Psychology},
    volume     = {62},
    number     = {1},
    pages      = {621--647},
    issn       = {0066-4308, 1545-2085},
    doi        = {10.1146/annurev.psych.093008.131123},
    urldate    = {2023-10-16},
    langid     = {english}
}

@inproceedings{haleProbabilisticEarleyParser2001,
    title     = {A {{Probabilistic Earley Parser}} as a {{Psycholinguistic Model}}},
    booktitle = {Second {{Meeting}} of the {{North American Chapter}} of the {{Association}} for {{Computational Linguistics}}},
    author    = {Hale, John},
    year      = 2001,
    urldate   = {2025-05-02}
}

@article{fischlerBrainPotentialsRelated1983,
    title    = {Brain Potentials Related to Stages of Sentence Verification},
    author   = {Fischler, I. and Bloom, P. A. and Childers, D. G. and Roucos, S. E. and Perry, N. W.},
    year     = 1983,
    month    = jul,
    journal  = {Psychophysiology},
    volume   = {20},
    number   = {4},
    pages    = {400--409},
    issn     = {0048-5772},
    doi      = {10.1111/j.1469-8986.1983.tb00920.x},
    langid   = {english},
    pmid     = {6356204}
}

@article{kriegerLimitsLLMSurprisal2024,
    title    = {On the Limits of {{LLM}} Surprisal as Functional {{Explanation}} of {{ERPs}}},
    author   = {Krieger, Benedict and Brouwer, Harm and Aurnhammer, Christoph and Crocker, Matthew W.},
    year     = 2024,
    journal  = {Proceedings of the Annual Meeting of the Cognitive Science Society},
    volume   = {46},
    number   = {0},
    urldate  = {2024-10-24},
    langid   = {english}
}

@article{kuperbergElectrophysiologicalDistinctionsProcessing2003,
    title    = {Electrophysiological Distinctions in Processing Conceptual Relationships within Simple Sentences},
    author   = {Kuperberg, Gina R. and Sitnikova, Tatiana and Caplan, David and Holcomb, Phillip J.},
    year     = 2003,
    month    = jun,
    journal  = {Brain Research. Cognitive Brain Research},
    volume   = {17},
    number   = {1},
    pages    = {117--129},
    issn     = {0926-6410},
    doi      = {10.1016/s0926-6410(03)00086-7},
    langid   = {english},
    pmid     = {12763198}
}

@article{frankWordPredictabilitySemantic2017,
    title     = {Word Predictability and Semantic Similarity Show Distinct Patterns of Brain Activity during Language Comprehension},
    author    = {Frank, Stefan L. and Willems, Roel M.},
    year      = 2017,
    month     = oct,
    journal   = {Language, Cognition and Neuroscience},
    volume    = {32},
    number    = {9},
    pages     = {1192--1203},
    publisher = {Routledge},
    issn      = {2327-3798},
    doi       = {10.1080/23273798.2017.1323109},
    urldate   = {2024-04-22}
}

@article{frankERPResponseAmount2015,
    title    = {The {{ERP}} Response to the Amount of Information Conveyed by Words in Sentences},
    author   = {Frank, Stefan L. and Otten, Leun J. and Galli, Giulia and Vigliocco, Gabriella},
    year     = 2015,
    month    = jan,
    journal  = {Brain and Language},
    volume   = {140},
    pages    = {1--11},
    issn     = {0093934X},
    doi      = {10.1016/j.bandl.2014.10.006},
    urldate  = {2023-11-30},
    langid   = {english}
}

@article{frankEyetrackingwithEEGCoregistrationCorpus2024,
    title    = {An Eye-Tracking-with-{{EEG}} Coregistration Corpus of Narrative Sentences},
    author   = {Frank, Stefan L. and Aumeistere, Anna},
    year     = 2024,
    month    = jun,
    journal  = {Language Resources and Evaluation},
    volume   = {58},
    number   = {2},
    pages    = {641--657},
    issn     = {1574-0218},
    doi      = {10.1007/s10579-023-09684-x},
    urldate  = {2024-11-28},
    langid   = {english}
}

@article{pimentelEffectAnticipationReading2023,
    title    = {On the {Effect} of {Anticipation} on {Reading} {Times}},
    volume   = {11},
    issn     = {2307-387X},
    url      = {https://doi.org/10.1162/tacl_a_00603},
    doi      = {10.1162/tacl_a_00603},
    urldate  = {2026-04-17},
    journal  = {Transactions of the Association for Computational Linguistics},
    author   = {Pimentel, Tiago and Meister, Clara and Wilcox, Ethan G. and Levy, Roger P. and Cotterell, Ryan},
    month    = dec,
    year     = {2023},
    pages    = {1624--1642}
}

@article{ehrlichContextualEffectsWord1981,
    title    = {Contextual Effects on Word Perception and Eye Movements during Reading},
    author   = {Ehrlich, Susan F. and Rayner, Keith},
    year     = 1981,
    month    = dec,
    journal  = {Journal of Verbal Learning and Verbal Behavior},
    volume   = {20},
    number   = {6},
    pages    = {641--655},
    issn     = {0022-5371},
    doi      = {10.1016/S0022-5371(81)90220-6},
    urldate  = {2023-12-20}
}

@article{shainWordFrequencyPredictability2024,
    title    = {Word {{Frequency}} and {{Predictability Dissociate}} in {{Naturalistic Reading}}},
    author   = {Shain, Cory},
    year     = 2024,
    month    = mar,
    journal  = {Open Mind},
    volume   = {8},
    pages    = {177--201},
    issn     = {2470-2986},
    doi      = {10.1162/opmi_a_00119},
    urldate  = {2025-03-25}
}

@article{stoneRoleSyntacticSemantic2025,
    title    = {The {{Role}} of {{Syntactic}} and {{Semantic Cues}} in {{Preventing Temporary Illusions}} of {{Plausibility}}},
    author   = {Stone, Kate and Rabovsky, Milena},
    year     = 2025,
    month    = sep,
    journal  = {Journal of Cognitive Neuroscience},
    volume   = {37},
    number   = {9},
    pages    = {1535--1561},
    issn     = {0898-929X},
    doi      = {10.1162/jocn_a_02320},
    urldate  = {2026-01-21}
}

@article{chowWaitSecondDelayed2018,
    title     = {Wait a Second! Delayed Impact of Argument Roles on on-Line Verb Prediction},
    author    = {Chow, Wing-Yee and Lau, Ellen and Wang, Suiping and Phillips, Colin},
    year      = 2018,
    month     = aug,
    journal   = {Language, Cognition and Neuroscience},
    volume    = {33},
    number    = {7},
    pages     = {803--828},
    publisher = {Routledge},
    issn      = {2327-3798},
    doi       = {10.1080/23273798.2018.1427878},
    urldate   = {2026-01-21}
}

@article{pynteOnlineContextualInfluences2008,
    title      = {On-Line Contextual Influences during Reading Normal Text: {{A}} Multiple-Regression Analysis},
    shorttitle = {On-Line Contextual Influences during Reading Normal Text},
    author     = {Pynte, Joel and New, Boris and Kennedy, Alan},
    year       = 2008,
    month      = sep,
    journal    = {Vision Research},
    volume     = {48},
    number     = {21},
    pages      = {2172--2183},
    issn       = {0042-6989},
    doi        = {10.1016/j.visres.2008.02.004},
    urldate    = {2026-01-21}
}

@inproceedings{mitchellSyntacticSemanticFactors2010,
    title      = {Syntactic and {{Semantic Factors}} in {{Processing Difficulty}}: {{An Integrated Measure}}},
    shorttitle = {Syntactic and {{Semantic Factors}} in {{Processing Difficulty}}},
    booktitle  = {Proceedings of the 48th {{Annual Meeting}} of the {{Association}} for {{Computational Linguistics}}},
    author     = {Mitchell, Jeff and Lapata, Mirella and Demberg, Vera and Keller, Frank},
    editor     = {Haji{\v c}, Jan and Carberry, Sandra and Clark, Stephen and Nivre, Joakim},
    year       = 2010,
    month      = jul,
    pages      = {196--206},
    publisher  = {Association for Computational Linguistics},
    address    = {Uppsala, Sweden},
    urldate    = {2026-01-21}
}

@article{traxlerPrimingSentenceProcessing2000,
    title      = {Priming in {{Sentence Processing}}: {{Intralexical Spreading Activation}}, {{Schemas}}, and {{Situation Models}}},
    shorttitle = {Priming in {{Sentence Processing}}},
    author     = {Traxler, Matthew J. and Foss, Donald J. and Seely, Rachel E. and Kaup, Barbara and Morris, Robin K.},
    year       = 2000,
    month      = nov,
    journal    = {Journal of Psycholinguistic Research},
    volume     = {29},
    number     = {6},
    pages      = {581--595},
    issn       = {1573-6555},
    doi        = {10.1023/A:1026416225168},
    urldate    = {2026-01-25},
    langid     = {english}
}

@incollection{nicenboimComputationalBayesianData2025,
    title     = {The influence of priors: sensitivity analysis},
    booktitle = {Introduction to {{Bayesian Data Analysis}} for {{Cognitive Science}}},
    author    = {Nicenboim, Bruno and Schad, Daniel J. and Vasishth, Shravan},
    year      = {2025},
    edition   = {1st edn},
    pages     = {634},
    chapter   = {3.4},
    publisher = {Chapman \& Hall},
    langid    = {english}
}

@article{wongPredictionReadingReview2024,
    title      = {Prediction in Reading: {{A}} Review of Predictability Effects, Their Theoretical Implications, and Beyond},
    shorttitle = {Prediction in Reading},
    author     = {Wong, Roslyn and Reichle, Erik D. and Veldre, Aaron},
    year       = 2024,
    month      = oct,
    journal    = {Psychonomic Bulletin \& Review},
    issn       = {1531-5320},
    doi        = {10.3758/s13423-024-02588-z},
    urldate    = {2025-03-12},
    langid     = {english}
}

@article{spacy,
    author    = {Honnibal, Matthew and Montani, Ines and Van Landeghem, Sofie and Boyd, Adriane},
    doi       = {10.5281/zenodo.1212303},
    title     = {{spaCy: Industrial-strength Natural Language Processing in Python}},
    year      = 2020
}

@inproceedings{yamada2020wikipedia2vec,
    title     = {{W}ikipedia2{V}ec: An Efficient Toolkit for Learning and Visualizing the Embeddings of Words and Entities from {W}ikipedia},
    author    = {Yamada, Ikuya and Asai, Akari and Sakuma, Jin and Shindo, Hiroyuki and Takeda, Hideaki and Takefuji, Yoshiyasu and Matsumoto, Yuji},
    booktitle = {Proceedings of the 2020 Conference on Empirical Methods in Natural Language Processing: System Demonstrations},
    year      = {2020},
    publisher = {Association for Computational Linguistics},
    pages     = {23--30},
    doi       = {10.18653/v1/2020.emnlp-demos.4}
}

@misc{banar2025mtebnle5nlembeddingbenchmark,
    title         = {MTEB-NL and E5-NL: Embedding Benchmark and Models for Dutch},
    author        = {Nikolay Banar and Ehsan Lotfi and Jens Van Nooten and Cristina Arhiliuc and Marija Kliocaite and Walter Daelemans},
    year          = {2025},
    eprint        = {2509.12340},
    archiveprefix = {arXiv},
    primaryclass  = {cs.CL},
    url           = {https://arxiv.org/abs/2509.12340}
}

@article{wang2024multilingual,
    title   = {Multilingual E5 Text Embeddings: A Technical Report},
    author  = {Wang, Liang and Yang, Nan and Huang, Xiaolong and Yang, Linjun and Majumder, Rangan and Wei, Furu},
    journal = {arXiv preprint arXiv:2402.05672},
    year    = {2024}
}

@article{brms,
    title    = {brms: An R Package for Bayesian Multilevel Models Using Stan},
    volume   = {80},
    url      = {https://www.jstatsoft.org/index.php/jss/article/view/v080i01},
    doi      = {10.18637/jss.v080.i01},
    number   = {1},
    journal  = {Journal of Statistical Software},
    author   = {Bürkner, Paul-Christian},
    year     = {2017},
    pages    = {1–28}
}

@manual{rcoreteam,
    title        = {R: A Language and Environment for Statistical Computing},
    author       = {{R Core Team}},
    organization = {R Foundation for Statistical Computing},
    address      = {Vienna, Austria},
    year         = {2024},
    url          = {https://www.R-project.org/}
}

@inproceedings{parvizUsingLanguageModels2011,
    title     = {Using {{Language Models}} and {{Latent Semantic Analysis}} to {{Characterise}} the {{N400m Neural Response}}},
    booktitle = {Proceedings of the {{Australasian Language Technology Association Workshop}} 2011},
    author    = {Parviz, Mehdi and Johnson, Mark and Johnson, Blake and Brock, Jon},
    editor    = {Molla, Diego and Martinez, David},
    year      = 2011,
    month     = dec,
    pages     = {38--46},
    address   = {Canberra, Australia},
    urldate   = {2026-02-02}
}

@misc{stan,
    title  = {Stan {{Reference Manual}}},
    author = {{Stan Development Team}},
    year   = 2023,
    url    = {https://mc-stan.org}
}

@article{dickeyWeightedLikelihoodRatio1970,
    title     = {The {{Weighted Likelihood Ratio}}, {{Sharp Hypotheses}} about {{Chances}}, the {{Order}} of a {{Markov Chain}}},
    author    = {Dickey, James M. and Lientz, B. P.},
    year      = 1970,
    month     = feb,
    journal   = {The Annals of Mathematical Statistics},
    volume    = {41},
    number    = {1},
    pages     = {214--226},
    publisher = {Institute of Mathematical Statistics},
    issn      = {0003-4851, 2168-8990},
    doi       = {10.1214/aoms/1177697203},
    urldate   = {2026-02-11}
}

@article{nicenboimStatisticalMethodsLinguistic2016,
    title      = {Statistical Methods for Linguistic Research: {{Foundational Ideas}}---{{Part II}}},
    shorttitle = {Statistical Methods for Linguistic Research},
    author     = {Nicenboim, Bruno and Vasishth, Shravan},
    year       = 2016,
    journal    = {Language and Linguistics Compass},
    volume     = {10},
    number     = {11},
    pages      = {591--613},
    issn       = {1749-818X},
    doi        = {10.1111/lnc3.12207},
    urldate    = {2026-02-11},
    langid     = {english}
}

@inproceedings{salicchiDifferentReadingProcessing2025,
    title      = {Different {{Reading Processing Stages}} or {{Different Brain Areas}}? {{A Computational Cognitive Investigation}} on {{N400}}, {{P600}}, and {{PNP}}},
    shorttitle = {Different {{Reading Processing Stages}} or {{Different Brain Areas}}?},
    booktitle  = {Computational {{Psycholinguistics Meeting}} 2025},
    author     = {Salicchi, Lavinia and Hsu, Yu-Yin},
    year       = 2025,
    month      = oct,
    urldate    = {2026-01-08},
    langid     = {english},
    note       = {Conference presentation abstract},
    url        = {https://openreview.net/forum?id=nu7Ld3AXWL}
}

@article{nieuwlandTestingLimitsSemantic2005,
    title      = {Testing the Limits of the Semantic Illusion Phenomenon: {{ERPs}} Reveal Temporary Semantic Change Deafness in Discourse Comprehension},
    shorttitle = {Testing the Limits of the Semantic Illusion Phenomenon},
    author     = {Nieuwland, Mante S. and Van Berkum, Jos J. A.},
    year       = 2005,
    month      = aug,
    journal    = {Cognitive Brain Research},
    volume     = {24},
    number     = {3},
    pages      = {691--701},
    issn       = {0926-6410},
    doi        = {10.1016/j.cogbrainres.2005.04.003},
    urldate    = {2026-02-12}
}

@article{aurnhammerP600ContinuousIndex2023,
    title     = {The {{P600}} as a Continuous Index of Integration Effort},
    author    = {Aurnhammer, Christoph and Delogu, Francesca and Brouwer, Harm and Crocker, Matthew W.},
    year      = 2023,
    journal   = {Psychophysiology},
    volume    = {60},
    number    = {9},
    pages     = {e14302},
    issn      = {1469-8986},
    doi       = {10.1111/psyp.14302},
    urldate   = {2024-11-01},
    copyright = {\copyright{} 2023 The Authors. Psychophysiology published by Wiley Periodicals LLC on behalf of Society for Psychophysiological Research.},
    langid    = {english}
}

@article{brouwerGettingRealSemantic2012,
    title      = {Getting Real about {{Semantic Illusions}}: {{Rethinking}} the Functional Role of the {{P600}} in Language Comprehension},
    shorttitle = {Getting Real about {{Semantic Illusions}}},
    author     = {Brouwer, Harm and Fitz, Hartmut and Hoeks, John},
    year       = 2012,
    month      = mar,
    journal    = {Brain Research},
    volume     = {1446},
    pages      = {127--143},
    issn       = {00068993},
    doi        = {10.1016/j.brainres.2012.01.055},
    urldate    = {2024-11-01},
    copyright  = {https://www.elsevier.com/tdm/userlicense/1.0/},
    langid     = {english}
}

@article{bulkesSemanticConstraintReading2020,
    title      = {Semantic Constraint, Reading Control, and the Granularity of Form-Based Expectations during Semantic Processing: {{Evidence}} from {{ERPs}}},
    shorttitle = {Semantic Constraint, Reading Control, and the Granularity of Form-Based Expectations during Semantic Processing},
    author     = {Bulkes, Nyssa Z. and Christianson, Kiel and Tanner, Darren},
    year       = 2020,
    month      = feb,
    journal    = {Neuropsychologia},
    volume     = {137},
    pages      = {107294},
    issn       = {00283932},
    doi        = {10.1016/j.neuropsychologia.2019.107294},
    urldate    = {2023-11-28},
    langid     = {english}
}

@article{lukeProvoCorpusLarge2018,
    title      = {The {{Provo Corpus}}: {{A}} Large Eye-Tracking Corpus with Predictability Norms},
    shorttitle = {The {{Provo Corpus}}},
    author     = {Luke, Steven G. and Christianson, Kiel},
    year       = 2018,
    month      = apr,
    journal    = {Behavior Research Methods},
    volume     = {50},
    number     = {2},
    pages      = {826--833},
    issn       = {1554-3528},
    doi        = {10.3758/s13428-017-0908-4},
    urldate    = {2023-11-28},
    langid     = {english}
}

@article{dambacherFrequencyPredictabilityEffects2006,
    title    = {Frequency and Predictability Effects on Event-Related Potentials during Reading},
    author   = {Dambacher, Michael and Kliegl, Reinhold and Hofmann, Markus and Jacobs, Arthur M.},
    year     = 2006,
    month    = apr,
    journal  = {Brain Research},
    volume   = {1084},
    number   = {1},
    pages    = {89--103},
    issn     = {0006-8993},
    doi      = {10.1016/j.brainres.2006.02.010},
    urldate  = {2025-02-05}
}

@misc{ostergaardCorpusJointEEG2025,
    title     = {A {{Corpus}} of {{Joint EEG}} and {{Self-Paced Reading}} of {{Natural Dutch Texts}}},
    author    = {Østergaard, Sara Møller and Lichtenberg, Lenneke and Boon, Laura and Nicenboim, Bruno},
    year      = 2025,
    month     = oct,
    number    = {g32rp\_v2},
    publisher = {PsyArXiv},
    doi       = {10.31234/osf.io/g32rp_v2},
    urldate   = {2026-02-19}
}

@inproceedings{gaoSimCSESimpleContrastive2021,
    title      = {{{SimCSE}}: {{Simple Contrastive Learning}} of {{Sentence Embeddings}}},
    shorttitle = {{{SimCSE}}},
    booktitle  = {Proceedings of the 2021 {{Conference}} on {{Empirical Methods}} in {{Natural Language Processing}}},
    author     = {Gao, Tianyu and Yao, Xingcheng and Chen, Danqi},
    editor     = {Moens, Marie-Francine and Huang, Xuanjing and Specia, Lucia and Yih, Scott Wen-tau},
    year       = 2021,
    month      = nov,
    pages      = {6894--6910},
    publisher  = {Association for Computational Linguistics},
    address    = {Online and Punta Cana, Dominican Republic},
    doi        = {10.18653/v1/2021.emnlp-main.552},
    urldate    = {2026-02-19}
}

@inproceedings{devlinBERTPretrainingDeep2019,
    title      = {{{BERT}}: {{Pre-training}} of {{Deep Bidirectional Transformers}} for {{Language Understanding}}},
    shorttitle = {{{BERT}}},
    booktitle  = {Proceedings of the 2019 {{Conference}} of the {{North American Chapter}} of the {{Association}} for {{Computational Linguistics}}: {{Human Language Technologies}}, {{Volume}} 1 ({{Long}} and {{Short Papers}})},
    author     = {Devlin, Jacob and Chang, Ming-Wei and Lee, Kenton and Toutanova, Kristina},
    editor     = {Burstein, Jill and Doran, Christy and Solorio, Thamar},
    year       = 2019,
    month      = jun,
    pages      = {4171--4186},
    publisher  = {Association for Computational Linguistics},
    address    = {Minneapolis, Minnesota},
    doi        = {10.18653/v1/N19-1423},
    urldate    = {2026-02-19}
}

@inproceedings{liTransformingGenericCoder2025,
    title     = {Transforming {{Generic Coder LLMs}} to {{Effective Binary Code Embedding Models}} for {{Similarity Detection}}},
    booktitle = {The {{Thirty-ninth Annual Conference}} on {{Neural Information Processing Systems}}},
    author    = {Li, Litao and Song, Leo and Ding, Steven and Fung, Benjamin C. M. and Charland, Philippe},
    year      = 2025,
    month     = oct,
    urldate   = {2026-02-20},
    langid    = {english}
}

@article{gpt,
    title    = {Improving {{Language Understanding}} by {{Generative Pre-Training}}},
    author   = {Radford, Alec and Narasimhan, Karthik and Salimans, Tim and Sutskever, Ilya},
    year     = 2018,
    journal  = {OpenAI Technical Report},
    langid   = {english}
}

@misc{llama,
    title         = {{{LLaMA}}: {{Open}} and {{Efficient Foundation Language Models}}},
    shorttitle    = {{{LLaMA}}},
    author        = {Touvron, Hugo and Lavril, Thibaut and Izacard, Gautier and Martinet, Xavier and Lachaux, Marie-Anne and Lacroix, Timoth{\'e}e and Rozi{\`e}re, Baptiste and Goyal, Naman and Hambro, Eric and Azhar, Faisal and Rodriguez, Aurelien and Joulin, Armand and Grave, Edouard and Lample, Guillaume},
    year          = 2023,
    month         = feb,
    number        = {arXiv:2302.13971},
    eprint        = {2302.13971},
    primaryclass  = {cs},
    publisher     = {arXiv},
    doi           = {10.48550/arXiv.2302.13971},
    urldate       = {2026-02-20},
    archiveprefix = {arXiv}
}

@article{enevoldsen2025mmtebmassivemultilingualtext,
    author    = {Kenneth Enevoldsen and Isaac Chung and Imene Kerboua and Márton Kardos and Ashwin Mathur and David Stap and Jay Gala and Wissam Siblini and Dominik Krzemiński and Genta Indra Winata and Saba Sturua and Saiteja Utpala and Mathieu Ciancone and Marion Schaeffer and Gabriel Sequeira and Diganta Misra and Shreeya Dhakal and Jonathan Rystrøm and Roman Solomatin and Ömer Çağatan and Akash Kundu and Martin Bernstorff and Shitao Xiao and Akshita Sukhlecha and Bhavish Pahwa and Rafał Poświata and Kranthi Kiran GV and Shawon Ashraf and Daniel Auras and Björn Plüster and Jan Philipp Harries and Loïc Magne and Isabelle Mohr and Mariya Hendriksen and Dawei Zhu and Hippolyte Gisserot-Boukhlef and Tom Aarsen and Jan Kostkan and Konrad Wojtasik and Taemin Lee and Marek Šuppa and Crystina Zhang and Roberta Rocca and Mohammed Hamdy and Andrianos Michail and John Yang and Manuel Faysse and Aleksei Vatolin and Nandan Thakur and Manan Dey and Dipam Vasani and Pranjal Chitale and Simone Tedeschi and Nguyen Tai and Artem Snegirev and Michael Günther and Mengzhou Xia and Weijia Shi and Xing Han Lù and Jordan Clive and Gayatri Krishnakumar and Anna Maksimova and Silvan Wehrli and Maria Tikhonova and Henil Panchal and Aleksandr Abramov and Malte Ostendorff and Zheng Liu and Simon Clematide and Lester James Miranda and Alena Fenogenova and Guangyu Song and Ruqiya Bin Safi and Wen-Ding Li and Alessia Borghini and Federico Cassano and Hongjin Su and Jimmy Lin and Howard Yen and Lasse Hansen and Sara Hooker and Chenghao Xiao and Vaibhav Adlakha and Orion Weller and Siva Reddy and Niklas Muennighoff},
    doi       = {10.48550/arXiv.2502.13595},
    journal   = {arXiv preprint arXiv:2502.13595},
    publisher = {arXiv},
    title     = {MMTEB: Massive Multilingual Text Embedding Benchmark},
    url       = {https://arxiv.org/abs/2502.13595},
    year      = {2025}
}

@inproceedings{penningtonGloVeGlobalVectors2014,
    title      = {{{GloVe}}: {{Global Vectors}} for {{Word Representation}}},
    shorttitle = {{{GloVe}}},
    booktitle  = {Proceedings of the 2014 {{Conference}} on {{Empirical Methods}} in {{Natural Language Processing}} ({{EMNLP}})},
    author     = {Pennington, Jeffrey and Socher, Richard and Manning, Christopher},
    editor     = {Moschitti, Alessandro and Pang, Bo and Daelemans, Walter},
    year       = 2014,
    month      = oct,
    pages      = {1532--1543},
    publisher  = {Association for Computational Linguistics},
    address    = {Doha, Qatar},
    doi        = {10.3115/v1/D14-1162},
    urldate    = {2026-02-23}
}

@misc{mikolovEfficientEstimationWord2013,
    title         = {Efficient {{Estimation}} of {{Word Representations}} in {{Vector Space}}},
    author        = {Mikolov, Tomas and Chen, Kai and Corrado, Greg and Dean, Jeffrey},
    year          = 2013,
    month         = sep,
    number        = {arXiv:1301.3781},
    eprint        = {1301.3781},
    primaryclass  = {cs},
    publisher     = {arXiv},
    doi           = {10.48550/arXiv.1301.3781},
    urldate       = {2026-02-23},
    archiveprefix = {arXiv}
}

@article{bojanowskiEnrichingWordVectors2017,
    title     = {Enriching {{Word Vectors}} with {{Subword Information}}},
    author    = {Bojanowski, Piotr and Grave, Edouard and Joulin, Armand and Mikolov, Tomas},
    editor    = {Lee, Lillian and Johnson, Mark and Toutanova, Kristina},
    year      = 2017,
    journal   = {Transactions of the Association for Computational Linguistics},
    volume    = {5},
    pages     = {135--146},
    publisher = {MIT Press},
    address   = {Cambridge, MA},
    doi       = {10.1162/tacl_a_00051},
    urldate   = {2026-02-23}
}

@misc{openai_gpt52_2025,
    author = {OpenAI},
    title  = {{GPT}-5.2 via {OpenAI API}},
    year   = {2025},
    url    = {https://developers.openai.com/api/docs/models/gpt-5.2}
}

\clearpage
\onecolumn
\appendix
\section{Appendices}
\subsection{Hugging Face References and Revision}
\begin{table*}[htb]
    \centering
    \begin{tabular}{p{0.5\textwidth}p{0.48\textwidth}}
    \hline
    \textbf{Hugging Face Reference} & \textbf{Revision}\\ \hline 
    Word2vec/wikipedia2vec\_nlwiki\_20180420\_300d \citep{yamada2020wikipedia2vec} & f7c83ecdf955a4f482a12517ca52a1f4b81e43cf  \\ 
    clips/e5-large-trm-nl \citep{banar2025mtebnle5nlembeddingbenchmark} & 683333f86ed9eb3699b5567f0fdabeb958d412b0   \\ 
    Word2vec/wikipedia2vec\_enwiki\_20180420\_100d \citep{yamada2020wikipedia2vec} & 7e4d6d224b95a5c351e2a47232701c4403ffbc16  \\ 
    fse/word2vec-google-news-300 &  528f381952a0b7d777bb4a611c4a43f588d48994  \\ 
    sentence-transformers/all-MiniLM-L6-v2 &  c9745ed1d9f207416be6d2e6f8de32d1f16199bf  \\ 
    intfloat/multilingual-e5-large \citep{wang2024multilingual} &  0dc5580a448e4284468b8909bae50fa925907bc5  \\ 
    spacy/nl\_core\_news\_sm \citep{spacy} & b9d28fe480eeacf9809fbd5ead5ef1ff27d9394e \\
    \hline
    \end{tabular}
    \caption{Overview of Hugging Face models used for the analysis.}
    \label{tab:hf-models}
\end{table*}

\subsection{Data loss}
The data loss for the different implementations of semantic association, i.e., the number of words in the corpus for which semantic association could not be calculated. The use of sentence embeddings resulted in a lower data loss compared to the word embedding implementations, as word embeddings extracted from a word2vec model only exist for a finite number of words. The data loss is largest for the \textit{CWE, Windowed(N=1)}.

\begin{table}[htb]
    \centering
    \begin{tabular}{lcc}
        \hline
        \textbf{Implementation} & \multicolumn{2}{c}{\textbf{Data loss}} \\
         & Overall & Content words \\
        \hline
        SE, All & 0.17\% & 0.08\% \\
        SE, Sentence(N=1) & 0.17\% & 0.08\% \\
        WE, All & 1.78\% & 2.06\% \\
        WE, Sentence(N=1) & 1.78\% & 2.06\% \\
        WE, Weighted & 1.78\% & 2.06\% \\
        CWE, All & 2.09\% & 2.33\% \\
        CWE, Sentence(N=1) & 2.17\% & 2.38\% \\
        CWE, Weighted & 2.09\% & 2.33\% \\
        CWE, Windowed(N=1) & 3.91\% & 4.25\% \\
        CWE, Windowed(N=2) & 2.21\% & 2.39\% \\
        \hline
    \end{tabular}
    \caption{Data loss caused by the different implementations of semantic association. The table shows the overall data loss across all words in the corpus, and the data loss for the current analysis that only considers content words.}
    \label{tab:data loss}
\end{table}

\subsection{Validation of semantic association}
\label{appendix:validation}
\begin{figure*}[htb]
    \centering
    \includegraphics[width=.95\textwidth]{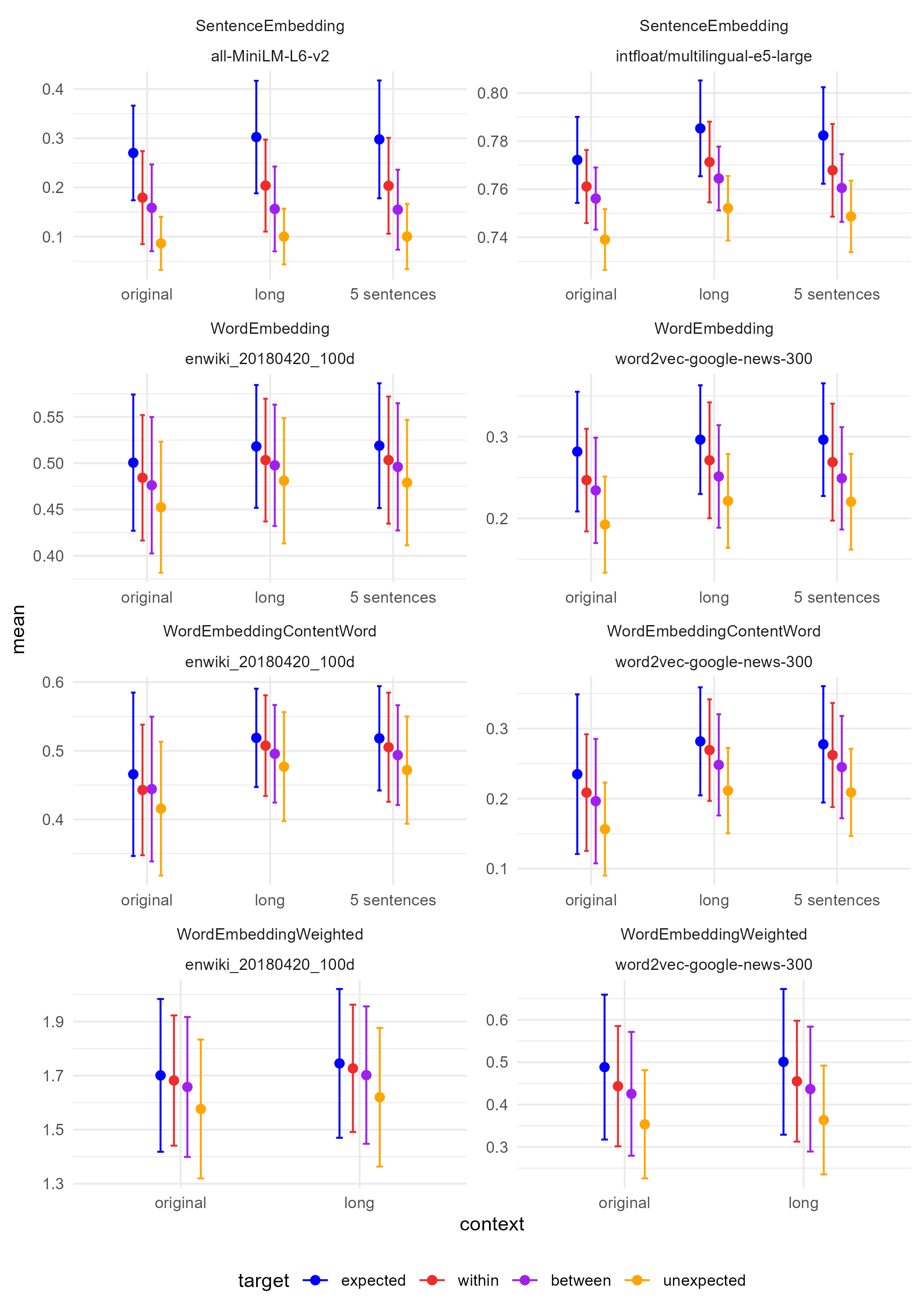}
    \caption{Average semantic association between context and target words (expected, within, between, and unexpected) from \citet{federmeierRoseAnyOther1999} given different implementations of semantic association. The plot is divided by the embedding model and the implementation of semantic association. The x-axis indicates the context the target was associated with, where ``long'' means the original and the generated longer context, ``original'' means the original context, ``5 sentences'' means the two sentences in the original context and three more from the longer context. The error bars indicate the standard deviation. Note that the y-axes are different across plots.}
    \label{fig:validation}
\end{figure*}

To validate the implementations of semantic association, we used data from \citet{federmeierRoseAnyOther1999}. As seen in \citet{ettingerModelingN400Amplitude2016}, we wanted to validate that the model identified expected targets as more semantically similar to the context compared to the within-category and between-category target words. In addition to the original stimuli, we added an unrelated target word (identical to an expected target in another context). Moreover, using OpenAI's GPT-5.2 \citep{openai_gpt52_2025}, we generated longer contexts of approx. 100 words to validate the models on contexts longer than the two-sentence context provided in the original data. The results of the validation are shown in Figure \ref{fig:validation}. All the implementations identify the correct ordering of semantic association between the target words and the contexts. The sentence embedding models, \texttt{all-MiniLM-L6-v2} and \texttt{intfloat/multilingual-e5-large}, generally exhibit less variance overlap between target words (and most notably between the expected and unexpected target) as compared to the word embedding models, \texttt{enwiki\_20180420\_100d} and \texttt{word2vec-google-news-300}.

\end{document}